\documentclass[10pt, conference, compsocconf]{IEEEtran}

\usepackage{color}
\usepackage{cite}
\usepackage[figure,vlined]{algorithm2e}
\SetCommentSty{textsf}
\SetKwRepeat{DoWhile}{do}{while}
\SetAlFnt{\small} 
\SetKw{Fn}{Procedure}
\newcommand{\Function}[2]{\Fn{\ProcNameSty{#1}}(#2)\par}
\usepackage{microtype}
\usepackage{bussproofs}
\usepackage{upgreek}
\usepackage{pmboxdraw}
\usepackage{ams}
\usepackage{pseudocode}
\usepackage{amssymb,amsmath,amsxtra,stmaryrd,verbatim,framed,wasysym}
\usepackage{epsfig,shadow,amssymb}
\usepackage{graphicx}
\usepackage{array}
\usepackage{latexsym}
\usepackage{subfigure}
\usepackage{listings}
\usepackage{paralist}
\usepackage{chngcntr}
\usepackage{amsfonts}
\usepackage{multirow}

\usepackage{amsthm}
\makeatletter
\def\th@plain{%
  \thm@notefont{}
\itshape 
}
\def\th@definition{%
  \thm@notefont{}
  \normalfont 
}
\makeatother

\newcommand{\commento}[1]{}
\newcommand{\algo}{\textsf{MagiCoder}}
\newcommand{\med}{\texttt{MedDRA}}
\newcommand{\vigi}{\texttt{VigiWork}}
\newcommand{\adr}{ADR}
\newcommand{\llt}{\texttt{LLT}}

\newcommand{\vllt}{\mathsf{Voted_{\llt}}}
\newcommand{\sllt}{\mathsf{Selected_{\llt}}}
\newcommand{\cone}[1]{\mathsf{C}_{1}(#1)}
\newcommand{\ctwo}[1]{\mathsf{C}_{2}(#1)}
\newcommand{\cthree}[1]{\mathsf{C}_{3}(#1)}
\newcommand{\cfour}[1]{\mathsf{C}_{4}(#1)}
\newcommand{\cfive}[1]{\mathsf{C}_{5}(#1)}

\newcommand{\voters}[1]{\mathsf{voters}_{#1}}
\newcommand{\voted}[1]{\mathsf{voted}_{#1}}
\newcommand{\weights}[1]{\mathsf{weights}_{#1}}
\newcommand{\svllt}{\mathsf{SortedVoted_{LLT}}}
\newcommand{\vigis}{\textsf{VigiSegn}}

\title{Automagically Encoding \\ Adverse Drug Reactions in MedDRA}

{
\author{
\IEEEauthorblockN{Margherita Zorzi, Carlo Combi} \IEEEauthorblockA{ 
 Department of Computer Science \\ University of Verona, Italy \\
\texttt{\scriptsize \{margherita.zorzi|carlo.combi\}@univr.it}
}
\and 
\IEEEauthorblockN{Riccardo Lora, Marco Pagliarini, Ugo Moretti} 
\IEEEauthorblockA{ Department of  Public Health and Community Medicine\\
 University of Verona, Italy\\
\texttt{\scriptsize \{riccardo.lora|marco.pagliarini|ugo.moretti\}@univr.it} 
}
}

\begin{document}

\maketitle
\begin{abstract}

Pharmacovigilance is the field of science devoted to the collection, analysis, and prevention of Adverse Drug Reactions (ADRs). Efficient strategies for the extraction of information about ADRs from free text sources are essential to support the important task of detecting and classifying unexpected pathologies, possibly related to (therapy-related) drug use. Narrative ADR descriptions may be collected in different ways, e.g., either by monitoring social networks or through the so called ``spontaneous reporting´´, the main method pharmacovigilance adopts in order to identify ADRs.
The encoding of free-text ADR descriptions according to MedDRA standard terminology is central for report analysis. It is a complex work, which has to be manually implemented by the pharmacovigilance experts. The manual encoding is expensive (in terms of time). Moreover, a problem about the accuracy of the encoding may occur, since the number of reports is growing up day by day.
In this paper, we propose $\algo$, an efficient Natural Language Processing algorithm able to automatically derive \med\ terminologies from free-text ADR descriptions. $\algo$  is part of \vigi, a web application for online ADR reporting and analysis.
From a practical point of view, $\algo$ reduces the encoding time of ADR reports.  Pharmacologists have simply to review and validate the \med\ terms proposed by $\algo$, instead of choosing the right terms among the 70K terms of \med. Such improvement in the efficiency of pharmacologists' work has a relevant impact also on the quality of the following data analysis. 

Our proposal is based on a general approach, not depending on the considered language. Indeed, we developed $\algo$ for the Italian pharmacovigilance language, but preliminarily analyses show that it is robust to language and dictionary changes. 

\end{abstract}


\section{Introduction}
\label{section-introduction}

Pharmacovigilance includes all activities aimed to systematically study risks and benefits related to the correct use of marketed drugs.
The development of a new drug, which begins with the production and ends with the commercialization of a drug, considers both pre-clinical studies (usually  tests on  animals) and clinical studies (tests on patients).
After these phases, a pharmaceutical company can require the authorization for the commercialization of the new drug. Notwithstanding, whereas at this stage drug benefits are well-know, results about drug safety are not conclusive~\cite{SafetyDrug}.
The pre-marketing tasks cited above have some limitations: they involve a small number of patients; they exclude relevant subgroups of population such as children and elders; the experimentation period is relatively short, less than two years; the experimentation does not deal with possibly concomitant pathologies, or with the concurrent use of other drugs.
For all these reasons, non-common Adverse Drug Reactions (ADRs), such as slowly-developing pathologies (e.g., carcinogenesis) or pathologies related to specific groups of patients, cannot be discovered before the commercialization. It may happen that drugs are withdrawn from the market after the detection of unexpected collateral effects.
Thus, it stands to reason that the control of ADRs is a necessity, considering the mass production of drugs. As a consequence, pharmacovigilance plays a crucial role in human healthcare improvement~\cite{SafetyDrug}. 

Spontaneous reporting is the main method pharmacovigilance adopts, in order to identify adverse drug reactions.
Through spontaneous reporting, health care professionals, patients, and pharmaceutical companies can voluntarily send information about suspected ADRs to the national regulatory authority~\footnote{in Italy, the Drug Italian Agency AIFA --Agenzia Italiana del FArmaco, http://www.agenziafarmaco.gov.it/}.
The spontaneous reporting is an important activity. It provides pharmacologists and regulatory authorites with early alerts, by considering every drug on the market and every patient category.

\begin{figure*} 
\begin{center}
\includegraphics[scale=.4]{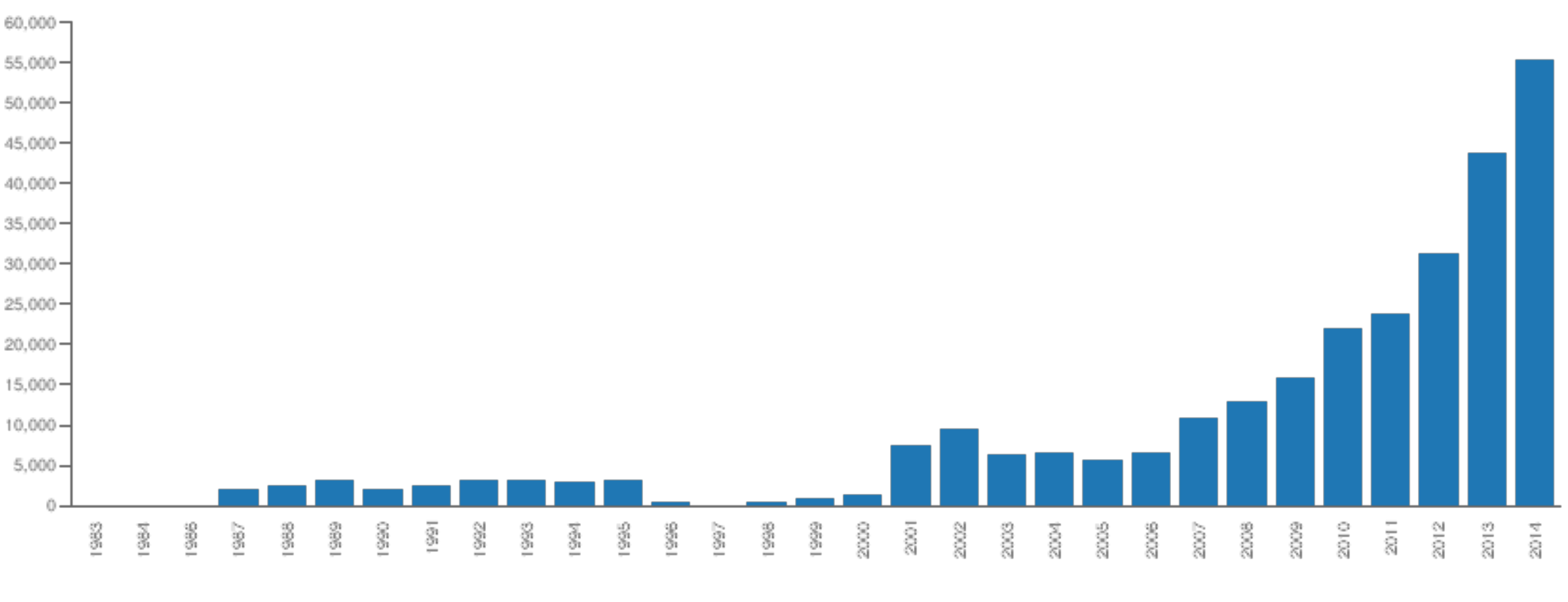}
\caption{The yearly increasing number of reports about suspected adverse reactions induced by drugs in Italy.}
\label{fig:Increasing-of-reports}
\end{center}
\end{figure*}

The Italian system of pharmacovigilance requires that in each local health structure there is a qualified person responsible for pharmacovigilance. Her/his assignment is to collect reports of suspected ADRs and to send them to the National Network of Pharmacovigilance (RNF) within seven days\footnote{According to the Italian Law, Art. 132 of Legislative Decree Number 219 of 04/24/2006.}.
Once reports have been notified and sent to RNF, currently through a web application, they are analysed by both local pharmacovigilance centres and by the Drug Italian Agency (AIFA). Subsequently, they are sent to Eudravigilance~\cite{Borg11} and to VigiBase~\cite{Aagard12} (the european and the worldwide pharmacovigilance network, RNF is part of, respectively).
In general, spontaneous ADR reports are filled by health care professionals (medical specialists, general practitioners, nurses, and so on), but also by citizens.
In the last years, Italian ADR reports have grown exponentially, going from approximately ten thousand in 2006 to around sixty thousand in 2014, as shown in Figure~\ref{fig:Increasing-of-reports}.

Since the post-marketing surveillance of drugs is of paramount importance, such an increase is certainly positive. At the same time, the manual review of reports became difficult and often unbearable both by people responsible for pharmacovigilance and by regional centres.
Indeed, each report must be checked, in order to control its quality; it is consequently encoded and transferred to RNF via ``copy by hand'' (actually, a printed copy).

Recently, to  increase the efficiency in collecting and managing ADR reports, a web application, called \vigi, has been designed and implemented for the Italian pharmacovigilance (at https://vigiwork.vigifarmaco.it/). 
Through \vigi, a spontaneous report can be inserted online both by healthcare professionals and by citizens (through different forms), as anonymous or registered users.
\vigi\ is user-friendly. The user is guided in compiling the report, since it has to be filled step-by-step (each phase corresponds to a different report section, i.e., ``Patient'', ``Adverse Drug Reaction'', ``Drug Treatments'' and ``Reporter'', respectively). Inserted data are then validated, since a report can be successfully sent only after completing the correct sequence of steps.

\vigi\ is also useful for pharmacovigilance supervisors. Indeed, \vigi\ reports are high-quality documents, since they are automatically validated (the presence, the format, and the consistency of data are validated at the filling time). As a consequence, they are easier to review (especially with respect to printed reports). Moreover, thanks to \vigi, a pharmacologist can send reports to RNF by simply pressing a button, after reviewing it.

Online reports have grown up to become the 30\% of the total number of Italian reports. As expected, it has been possible to observe that the average time between the dispatch of online reports and the insertion into RNF is sensibly shorter with respect to the the insertion from printed reports.
Notwithstanding, there is an operation which still requires the manual work of people responsible for Pharmacovigilance also for online report revisions: the encoding in \med\ terminology of the free text, through which the reporter describes one or more  adverse drug reactions.
The description of a suspected ADR through narrative text could seem redundant/useless. Indeed, one could reasonably imagine sound solutions based either on an autocompletion form or on a menu with \med\ terms. In these solutions, the description of ADRs would be directly encoded by the reporter and no expert work for \med\ terminology extraction would be required. 
However, such solutions are not completely suited for the pharmacovigilance domain and the narrative description of ADRs remains a desirable feature, for at least two reasons. First, the description of an ADR by means of one of the seventy thousand \med\ terms is  a complex task. In most cases, the reporter which points out the adverse reaction is not an expert in \med\ terminology. This holds in particular for citizens, but it is still valid for several professionals. Thus, describing ADRs  by means of natural language sentences is simpler.
Second, the choice of the suitable term(s) from a given list or from an autocompletion field can influence the reporter and limit her/his expressiveness. As a consequence, the \emph{quality} of the description would be also in this case undermined.
Therefore, \vigi\ offers a free-text form for specifying and ADR with all the possible details, without any restriction about the content or limits to the length of the written text. Consequently, \med\ encoding has then to be manually implemented by qualified people responsible for pharmacovigilance, before the transmission to RNF. As this work is expensive in terms of time and attention required, a problem about the accuracy of the encoding may occur given the continuous growing of the number of reports. 

According to the described scenario, in this paper we propose $\algo$, a natural language processing (NLP)~\cite{Jurafsky2000} algorithm, which automatically assigns one or more \med\ term codes to each narrative ADR description in the online reports collected by \vigi.

The paper is organized as follows. In Section~\ref{sec:related} we provide some background notions and we discuss related work. 
In Section~\ref{sec:magicoder} we present the algorithm \algo, by providing both a qualitative description and the pseudocode. In Section~\ref{sec:uiandbenchmark} we spend some words about the user interface, we explain the benchmark we developed to test $\algo$ performances and we discuss first results.  Finally, in Section~\ref{sec:future} we discuss the main features of our work and sketch some future research lines.

\section{Background and Related Work}\label{sec:related}

\subsection{Natural Language Processing and Text Mining in Medicine}

Automatic detection of adverse drug reactions from text  recently received an increasing interest in pharmacovigilance research. 
Narrative descriptions of ADRs come from heterogeneous sources: spontaneous reporting, Electronic Health Records, Clinical Reports, and social media.  
In~\cite{Bate,Wang09,Friedman,Aramaki10,Dunagan08} some NLP approaches have been proposed for the extraction of ADRs from text. 
In~\cite{Bailey09}, the authors collect narrative discharge summaries from the Clinical Information System at New York Presbyterian Hospital. MedLEE, an NLP system, is applied to this collection, to identify medication events and entities, which could be potential adverse drug events. Co-occurrence statistics with adjusted volume tests were used to detect associations between the two types of entities, to calculate the strengths of the associations, and to determine their cutoff thresholds. 
In~\cite{Toldo12},  the authors report on the adaptation of a machine learning-based system for the identification and extraction of ADRs in case reports.
The role of NLP approaches in optimised machine learning algorithms is also explored in~\cite{Gonzalez14}, where the authors address the problem of  automatic detection of ADR assertive text segments from distinct sources, focusing on data posted by users on social media (Twitter and DailyStrenght, a health care oriented social media).  Existing methodologies for NLP are discussed; an experimental comparison between NLP-based machine learning algorithms over data sets from different sources has been proposed. Moreover, the authors address the issue of data imbalance for ADR description task. In~\cite{Yang12} the authors propose to use association mining and Proportional Reporting Ratios (PRR, a well-know pharmacovigilance statistical index) to mine the associations between drugs and adverse reactions from the user contributed content in social media. In order to extract adverse reactions from on line text (from health care communities), the authors apply the Consumer Health Vocabulary (at http://www.consumerhealthvocab.org) to generate ADR lexicon. ADR lexicon is a computerized collection of health expressions derived from actual consumer utterances (authored by consumers), linked to professional concepts and reviewed and validated by professionals and consumers. Narrative text is preprocessed following standard NLP techniques (such as stop word removal, see Section~\ref{sec:description}). An experiment using ten drugs and five adverse drug reactions is proposed. The Food and Drug Administration alerts are used as the gold standard, to test the performance of the proposed techniques. 
The authors developed algorithms to identify ADRs from threads of drugs, and implemented association mining to calculate leverage and lift for each possible pair of drugs and adverse reactions in the dataset. At the same time, PRR is also calculated.

Other interesting papers about pharmacovigilance and machine learning or data mining are, e.g.,~\cite{Harpaz10} and~\cite{Bellazzi15}. 
In~\cite{Uppsala} a text extraction tool is implemented on the .NET platform with functionalities for preprocessing text (removal of stop words, Porter stemming and use of synonyms) and matching medical terms using permutations of words and spelling variations (Soundex, Levenshtein distance and Longest common subsequence distance~\cite{Collins03}). Its performance has been evaluated on both manually extracted medical terms from summaries of product characteristics and unstructured adverse effect texts from Martindale (i.e. a medical reference for information about drugs and medicines) using the WHO-ART and \med\ medical term dictionaries. A lot of linguistic features have been considered and a careful analysis of performances has been provided.

\subsection{\med\ Dictionary}

The Medical Dictionary for Regulatory Activities (\med) is a medical terminology used to classify adverse event information associated with the use of biopharmaceuticals and other medical products (e.g., medical devices and vaccines). Coding these data to a standard set of MedDRA terms allows health authorities and the biopharmaceutical industry to exchange and analyze data related to the safe use of medical products~\cite{Radhakrishna14}. It has been developed by the International Conference on Harmonization (ICH); it belongs to the International Federation of Pharmaceutical Manufacturers and Associations (IFPMA); it is controlled and periodically revised by the \med\ Mainteinance And Service Organization (MSSO).
\med\ is available for eleven European languages and for Chinese and Japanese too. It is updated twice a year (in March and in September), following a collaboration-based approach: everyone can propose new reasonable updates or changes (as effects of events as the onset of new pathologies) and a team of experts eventually decides about the publication of the updates.
MedDRA terms are organised into a hierarchy: 
the SOC (System Organ Classes) level includes the most general terms; 
the LLT (Low Level Terms) level includes more specific terminologies; 
between SOC and LLT there are three intermediate levels (HLGT, HLT and PT).

 Table~\ref{tab:exmeddra} shows an example of the hierarchy: the reaction \emph{Itch} is described starting from \emph{Skin disorders}, \emph{Epidermal conditions}, \emph{Dermatitis and Eczem}, and \emph{Asteatotic Eczema}. Preferred Terms are Low Level Terms chosen to be the representative of a group of terms. It should be stressed that the hierarchy is multiaxial: for example, a PT (Preferred Term) can be grouped in one or more HLT (High Level Term), but it belongs to only one primary SOC (System Organ Class) term.
 
 \begin{table}
  \begin{center}
    \begin{tabular}{| c| c |}
      \hline
      \med\ Level & \med\ Term     \\
      \hline
      &\vspace{-.2cm}\\
      SOC & Skin disorders           \\
           \hline
      &\vspace{-.2cm}\\
      HLGT & Epidermal conditions\\
           \hline
      &\vspace{-.2cm}\\
     HLT  & Dermatitis and Eczema \\
          \hline
      &\vspace{-.2cm}\\
      PT  & Asteatotic Eczema\\
           \hline
      &\vspace{-.2cm}\\
      LLT  & Itch\\
           \hline
    \end{tabular}
  \end{center}
\caption{MedDRA Hierarchy - an Example}\label{tab:exmeddra}
\end{table}

The encoding of ADRs through \med\ is extremely important for report analysis as for a prompt detection of problems related to drug-based treatments.
Thanks to \med\, it is possible to group similar/analogous cases described in different ways (e.g. by synonyms) or with different details/levels of abstraction.

\section{\algo: an Algorithm for ADR Automatic Encoding}\label{sec:magicoder}

 A natural language ADR description is a completely free text. The user has no limitations, she/he can potentially write everything: a number of online ADR descriptions actually contain information not directly related to drug effects.
An NLP software has to face and solve many issues: trivial orthographical errors; the use of singular versus plural nouns; the so called ``false positives''  i.e. syntactically retrieved inappropriate results, which are closely resembling correct solutions; the structure of the sentence, i.e. the way an assertion is built up in a given language.
Also the ``intelligent'' detection of linguistic connectives is a crucial issue. For example, the presence of a negation can potentially change the overall meaning of a description.

In general, a satisfactory automatization of human reasoning and work is a subtle task, and the uncontrolled extension of the dictionary with auxiliary synonymous or the naive ad-hoc management of particular cases can limit the efficiency of the algorithm.
For these reasons, we carefully designed $\algo$, even through a side-by-side collaboration between pharmacologists and computer scientists, in order to yield an efficient tool, capable to really support pharmacovigilance activities.

In literature, several NLP algorithms still exists, and several interesting approaches (such as the so called morpho-analysis of natural language) have been studied and proposed~\cite{BauerLaurie2003,Jurafsky2000,Kishida2005}. According to the described pharmacovigilance domain, we considered algorithms for the morpho-analysis and the part-of-speech extraction techniques~\cite{BauerLaurie2003,Jurafsky2000} too powerful and general purpose for the first solution to our problem. 

Thus, we decided to design and develop an ad-hoc algorithm for the problem we are facing, namely that of deriving \med\ terms from narrative text and mapping segments of text in effective \llt\ terms. This task has to be done in a very feasible time (we want that each interaction user/\algo\ requires less than a second) and the solution offered to the expert has to be readable and useful. Therefore, we decided to ignore the \emph{structure} of the narrative description and address the issue in a simpler way.
Main features of \algo\ can be summarized as follows:
\begin{itemize}
\item it requires \emph{a single linear scan of the narrative description}: as a consequence, our solution is particularly efficient in terms of computational complexity;
\item it has been designed and developed for the specific problem of mapping Italian text to \med\ dictionary, but we claim the way \algo\ has been developed is sound with respect to Language and Dictionary changes. 
\item the current version of \algo\ is only based on the pure syntactical recognition of the text and it does not exploit any heuristic or external synonym dictionary; as it will be discussed in Section~\ref{sec:uiandbenchmark}, experimental results are encouraging and  we empirically observed that the  use of an external dictionary produces a relevant improvement of performances.
\end{itemize}

\subsection{\algo: Overview}\label{sec:description}

The  main idea of $\algo$ is that a single linear scan of the free-text is sufficient, in order  to recognize $\med$ terms. 

From an abstract point of view, we try to recognize, in the narrative description, \emph{single words} belonging to \llt terms, which do not necessarily occupy consecutive positions in the description. This way, we try to \emph{reconstruct} \med\ terms, taking into account the fact that in a description the reporter can permute or omit words.
As we will show, \algo\ has not to deal with computationally expensive tasks, such as taking into account subroutines for permutations and combinations of words (as, for example, in~\cite{Uppsala}). 

We can distinguish  five phases in the procedure, we will discuss in detail in the following:
\begin{enumerate}
\item Preprocessing of the original text;
\item{Definition of ad-hoc data structures;}
\item Word-by-word linear scan of the description and ``voting task'';
\item Weight calculation and sorting of voted terms;
\item Winning terms release.
\end{enumerate}

\subsubsection{Preprocessing of the original \adr\ description}

Given a natural language \adr\ description, the text has to be preprocessed in order to perform an efficient computation. We adopt well-know techniques such as tokenization~\cite{Clark2010}, where a phrase is reduced to \emph{tokens}, i.e. syntactical units, which often, as in our case, correspond to words. A tokenized text can be easily manipulated as an enumerable object, e.g. an array.
A \emph{stop word} is a word which can be considered irrelevant for the text analysis (e.g. an article or an interjection). 
In this first release of our software we decided to not take into account \emph{connectives}, e.g. conjunctions, disjunctions, negations.
Once one has defined the set of the stop words, the original text is cleaned from such irrelevant words.

A fruitful preliminary work is  the extraction of the corresponding \emph{stemmed} version from the original tokenized (and stop-word free) text. 
Stemming is a linguistic technique that, given a word, reduces it to a particular kind of root form~\cite{Clark2010}. It is useful in text analysis, in order to avoid problems such as bad word recognition due to singular/plural forms (e.g., hand/hands). Stemming is also potentially harmful, since it can generate the so called ``false positives'' terms. A meaningful example can be found in Italian language. The  plural of the word \emph{mano} (in English, \emph{hand}) is \emph{mani} (in English, \emph{hands}), and their stemmed radix is \emph{man}, which is also the stemmed version of \emph{mania} (in English, \emph{mania}).

\subsubsection{Definition of ad hoc data structures}

The algorithm proceeds with a word-by-word comparison. We iterate on the preprocessed text and we test if a single word $w$ (a token) occurs into one or many \llt\ terms.

In order to  efficiently test if a token belongs to one or more \llt\ terms, we need to create a further level of the \med\ dictionary. 
{The \llt\ level of \med\ is actually a set of \emph{phrases}, i.e. \emph{sequences of words}. By scanning these sequences, we built a \emph{meta-dictionary} of all the words which compose  \llt\ terms.}  
As we will describe in Section~\ref{sec:pseudocode}, in $O(mk)$ time units (where $m$ and $k$ are the cardinality of the set of \llt\ terms and the length of the longest \llt\ term in \med, respectively) we can build a hash table having all the words occurring in \med\ as keys, where the value associated to key $w_i$ contains information about the set of  \llt s containing $w_i$.
This way, we can verify the presence in \med\ of a word $w$ encountered in the \adr\ description in constant time.  We call this meta-dictionary $\mathsf{DictByWord}$.
We build a meta dictionary also from a stemmed version of \med, to verify the presence of stemmed descriptions. We call it $\mathsf{DictByWordStem}$.

Also the \med\ dictionary is loaded for the computation into  hash tables and, in general, all our main data structures are dictionaries. We aim to stress that,  to retain efficiency, we preferred exact string matching with respect to approximate string matching, when looking for a word into the meta dictionary.  Approximate string matching would allow us to retrieve terms that would be lost in exact string matching (e.g., we could recognize misspelled words in the ADR description), but it would worsen the performances of the text recognition tool, since direct access to the dictionary would not be possible. We discuss the problem of addressing orthographical errors in Section~\ref{sec:future}. 

\subsubsection{Word-by-word linear scan of the description and voting task}\label{voting}
 
Algorithm $\algo$ scans the text word-by-word (remember that each word corresponds to a token) one time and performs a ``voting task'': at the $i$-th step,  it marks (i.e. ``votes''),  with index $i$ each \llt\ term $t$ containing the current (i-th) word of the \adr\ description.  Moreover, it keeps track of the position where the $i$-th word occurs in \llt\ terms.
$\algo$ tries to find a word match both for the exact and the stemmed version of the meta dictionary and keeps track of the kind of match it has eventually found. It updates a flag, initially set to 0, if at least a stemmed matching is found. If a word $w$ has been exactly recognized in a term $t$, the  match between the stemmed versions of $w$ and $t$ is not considered.
At the end of the scan, the procedure has built a sub-dictionary containing only terms ``voted'' at least by a word. We will call $\vllt$ the sub dictionary of voted terms. 

Each selected term $t$  is equipped with two auxiliary data structures, containing,  respectively:

\begin{enumerate}
\item the positions of the voting words in the \adr\ description; we will call $\voters{t}$ this sequence of indexes;
\item the positions of the voted words in the \med\ term $t$; we will call $\voted{t}$ this sequence of indexes.
\end{enumerate}

Moreover, we endow each selected term with a third structure that will contain the sorting criteria we define below; we will call it $\weights{t}$. 

Let us now introduce some notations we will use in the following. We denote as $t.size$ the function that, given a \llt\ term $t$, returns the number of words contained in $t$. We denote as $\voters{t}.length$ (resp. $\voted{t}.length$) the function that returns the number of indexes belonging to $\voters{t}$ (resp. $\voted{t}$). 
We denote as $\voters{t}.min$  and $\voters{t}.max$ the functions that return the maximum and the minimum indexes in  $\voters{t}$, respectively.

\subsubsection{Weight calculation and sorting}\label{sec:criteriasorting}

After the voting task, selected terms have to be ordered. Notice that a purely syntactical recognition of words in \llt\ terms potentially generates a large number of voted terms. So we have to: i) filter a subset of highly feasible solutions; ii) choose a good final selection criteria (this will be discuss in Section~\ref{sec:winning}).

To this end, we define five criteria as ``weights'' to assign to voted terms. In the following, $\frac{1}{t.size}$ is a normalization factor (w.r.t. the length, in terms of words, of the \llt\ term $t$). For the first four criteria the optimum value is 0.

\begin{description}
\item [\textbf{Criterion one: Coverage}]\mbox{}

First, we consider how many words of each voted \llt\ term have been recognized.

\begin{displaymath}
\cone{t}=\frac{t.size-\voters{t}.length}{t.size}
\end{displaymath}

\item [\textbf{Criterion two: Type of Coverage}]\mbox{}

The algorithm considers whether a perfect matching has been performed using or not stemmed words. $\ctwo{\cdot}$ is simply a flag. $\ctwo{t}$ holds if stemming has been used at least once in the voting procedure.

\item [\textbf{Criterion three: Coverage Distance}]\mbox{}

The use of stemming allows one to find a number of (otherwise lost) matches. As side effects, we often obtain a quite large set of joint winner candidate terms. In this phase, we introduce a string distance comparison between recognized words in the original text and retrieved \llt\ terms. Among the possible string metrics, we use the so called pair distance~\cite{string}, which is robust with respect to word permutation. So,

\begin{displaymath}
\cthree{t}=\mathit{pair}(t,\overline{t})
\end{displaymath}

where $\mathit{pair}(s,r)$ is the pair distance function (between strings $s$ and $r$) and $\overline{t}$ is the term ``rebuilt'' from the words in \adr\ description corresponding to indexes in $\voters{t}$.

\item [\textbf{Criterion four: Coverage Density}]\mbox{}

We want to estimate how an \llt\ term has been covered. 

\begin{displaymath}
\cfour{t}=\frac{(\voters{t}.max-\voters{t}.min)+1}{t.size}
\end{displaymath}

The intuitive meaning of the criterion is to quantify the ``quality'' of the coverage. If an \llt\ term has been covered by nearby words, it will be considered a good candidate for the solution. This Criterion has to be carefully implemented, taking into account possible duplicate words.

\item [\textbf{Criterion Five: Coverage Distribution}]\mbox{}

After the evaluation and the sorting by the criteria described above, good solutions are sorted in the first positions. We add a further criterion, the only one based on the assumptions we made about the structure of  (Italian) sentences. 
The following formula simply sums the index of the covered words for  $t\in\vllt$:

\begin{displaymath}
\cfive{t}=\sum_{i=0}^{\voted{t}.length-1} \voted{t}[i]
\end{displaymath}

If $\cfive{t}$ is small, it means that words in the first positions of term $t$ have been covered. We introduce this criterion to discriminate between possibly joint winning terms. Indeed, an Italian medical description of a pathology has frequently the following shape: \emph{name of the pathology}+\emph{``location'' or adjective}. Intuitively, we privilege terms, for which the recognized word(s) are probably the one(s) describing the pathology.


After computing (and storing) the weights related to the above criteria, for each voted term $t$ we have the structure $\weights{t}=[\cone{t}, \ctwo{t}, \cthree{t}, \cfour{t}, \cfive{t}]$, containing the weights corresponding to the five criteria.

We finally proceed by ordering voted terms by multiple value sorting (on elements in $\weights{t}$, $t\in\vllt$) and call $\svllt$ the sorted dictionary.

\end{description}

\subsubsection{Release of winning terms}\label{sec:winning}

In order to provide an effective support to pharmacovigilance experts' work, it is important to offer, among the ``good'' solutions of the algorithm (well positioned \llt\ terms in sorted output), a small subset of candidate solutions, typically from one to six terms recognized as the best match of the ADR description. We will call $\sllt$ such a set.
This is a subtle task. As previously said, the pure syntactical recognition of \med\ terms into a free-text generates a possibly large set of syntactically good results. 
Therefore, the releasing strategy has to be carefully designed. 
The main idea is to select and return a subset of voted terms which ``covers'' the \adr\ description.
We iterate on the ordered dictionary and for each  $t \in \svllt$ we iterate on $\voters{t}$ and we select $t$ if the following conditions hold: 1) $t$ does not belong to $\sllt$; 2) $t$ is not a prefix of another selected term $t' \in \vllt$; 3) for any $w_i \in \voters{t}$,  
$w_i$ has not been covered or $w_i$ has not been exactly covered (only the stemmed version has been eventually recognized) or $t$ has been ``voted'' without stemming.
\footnote{In the implementation we add also the following thresholds: we choose only terms $t$ such that $\cthree{t}<0.5$ and $\cfive{t}<3$. We extracted these threshold by means of some empirical tests. We plain to eventually adjust them after some further performance tests}.
We keep track of the words of the ADR description covered by the selection. We consider  all the sorted dictionary $\svllt$, but the selection actually ends when all the words of the description have been covered. The user interface (UI) of \vigi\ (described in Section~\ref{sec:uiandbenchmark}) further filters winning terms, by releasing from zero up to six solutions. 

In $\algo$ we do not need to consider ad hoc subroutines to address permutations and combinations of words (as it is done, for example, in~\cite{Uppsala}). In Natural Language Processing, permutations and combinations of words are important, since in spoken language the order of words can change w.r.t. the formal structure of the sentences. Moreover, some words can be omitted, while the sentence still retains the same meaning.  These aspects come for free from our voting procedure: after the scan, we retrieve the information that \emph{a set of words covers a term} $t\in\vllt$, but \emph{the order between words does not matter}.

\subsection{\algo: the Algorithm}\label{sec:pseudocode}

Figure~\ref{tab:pseudocode} depicts the pseudocode of \algo. Here we provide a high-level description  of the procedure. We represent dictionaries  either as sets of words or as sets of functions. As usually, the formula $w \in \mathsf{LLTDict}$ means ``word $w$ belongs to dictionary $\mathsf{DictByWord}$´´ (similarly for $\mathsf{DictByWordStem}$, $\vllt$, $\svllt$, $\sllt$).
Procedure $Preprocessing$ takes the narrative description, puts it into an array of words and performs tokenization and stop-word removal. 
Procedures $CreateMetaDict$ and $CreateMetaDictStem$ get the dictionary of \llt\ terms and create a dictionary of \emph{words} and of their stemmed versions, respectively, which belong to \llt\ terms, retaining the information about the set of terms containing each word.
By the functional notation $\mathsf{DictByWord}(j)$ (similarly, $\mathsf{DictByWordStem}(j)$),  we refer to the set of \llt\ terms containing the word $j$ (or its stemmed version).
Function $stem(i)$ returns the stemmed version of word $i$. 
Function $indx_{t}(j)$ returns the position of word $j$ in term $t$. $stem\_usage_{t}$ is a flag, initially set to 0, which holds 1 if at least a stemmed matching  with the \med\ term $t$ is found.
$\mathsf{adr\_clear}$, $\voters{t}$, $\voted{t}$ are arrays and  $\mathsf{add}[A,l]$ denotes the insertion of $l$ in array $A$, where $l$ is an element or a sequence of elements. 

$\mathsf{C}_{i}$ ($i=1,\ldots,5$) are criteria defined in Section~\ref{sec:criteriasorting} and procedure $sortby(v_1,\ldots,v_k)$ performs the multi-value sorting of values $v_1,\ldots,v_k$.
Procedure $prefix(S,t)$, where $S$ is a set of terms and $t$ is a term, tests whether $t$ (considered as a string) is \emph{prefix} of a term in $S$. Dually,  procedure $remove\_prefix(S,t)$  tests if in $S$ there are one or more prefixes of $t$, and eventually remove them from $S$.
Function $mark(j)$ specifies whether a word $j$ has been already covered in the (partial) solution during the term release:  $mark(j)$ holds 1 if $j$ has been covered (with or without stemming) and it holds 0 otherwise. We assume that before starting the final phase of building the solution (i.e., the returned set of $\llt$ terms), $mark(j)=0$ for any word $j$ belonging to the description.

\begin{algorithm*}[!t]
    \Function{\algo}{$D$ description, $\mathsf{LLTDict}$ dictionary}
    \KwIn{$D$: the narrative description; $\mathsf{LLTDict}$: a data structure 		         containing the $\llt$ terms of \med\ dictionary}
    \KwOut{a set of $\llt$ ordered terms}
     $\mathsf{DictByWord}$ = CreateMetaDict($\mathsf{LLTDict}$)\;
     $\mathsf{DictByWordStem}$ = CreateStemMetaDict($\mathsf{LLTDict}$)\;
     \emph{adr\_clear} = Preprocessing($D$)\;
     \emph{adr\_length} = \emph{adr\_clear}.length\;
    \ForEach{(i $\in [0, adr\_length-1]$ }{
    \tcc{test whether the current word belongs to  \med }
    		\If{adr\_clear[i] $\in \mathsf{DictByWord}$}{
        \tcc{for each term containing the word}
   			\ForEach{(t $\in \mathsf{DictByWord}$(adr\_clear[i])}{
    			\tcc{keep track of the index of the voting word}
   			$\mathsf{add}$[$\voters{t}$,i]\;
           \tcc{keep track of the index of the recognized word in $t$ }
			$\mathsf{add}$[$\voted{t}$, $indx_{t}$(\textit{adr\_clear[i]})]\;

  			$\vllt$ = $\vllt$ $\cup t$\;
   			}
   		}
      	\tcc{test if the current (stemmed) word belongs  the stemmed \med }
      	\If{stem({adr\_clear[i]}) $\in \mathsf{DictByWordStem}$}{
      		\ForEach{t $\in\mathsf{DictByWordStem}$(stem(adr\_clear[i]))}{
      			\tcc{ test if the current term has not been exactly voted by the same word }
 				\If{i $\notin \voters{t}$}{
 					$\mathsf{add}$[$\voters{t}$, i]\;
  					$\mathsf{add}$[$\voted{t}$, $indx_{t}$(\textit{adr\_clear[i]})]\;
     				\tcc{  keep track that $t$ has been covered by a stemmed word}
 					$stem\_usage_{t}$ = true\;
 				} 
  			$\vllt$ = $\vllt$ $\cup$ t
  			}
 		}
	}
	\tcc{ for each voted term, calculate the five weights of the corresponding criteria}
	\ForEach{t $\in\vllt$}{
		$\mathsf{add}$[$\weights{t}, \cone{t},\ctwo{t},\cthree{t},\cfour{t},\cfive{t}$]
	}  
	\tcc{  multiple value sorting of the voted terms}
	$\svllt = \vllt$.sortby($\mathsf{C}_{1},\mathsf{C}_{2},\mathsf{C}_{3},\mathsf{C}_{4},\mathsf{C}_{5}$)\;
	\ForEach{t $\in\svllt$}{
		\ForEach{index $\in\voters{t}$}{
			\tcc{select a term $t$ if its i-th voting word has not been covered or if its i-th voting word has been perfectly recognized in $t$ and if $t$ is not prefix of 
			another already selected terms}
			\If{(($stem\_usage_{t}$ = false OR (mark(adr\_clear(index))=0))
			AND t $\notin\sllt$ AND  prefix($\sllt$,t)=false)}{
				mark(adr\_clear(index))=1\;
				\tcc{remove from the selected term set all terms which are prefix of $t$}
				$\sllt$ = remove\_prefix($\sllt$,t)\;
				$\sllt$ = $\sllt \cup$ t	
			} 
		}
	}  
	\Return{$\sllt$}

	\caption{Pseudocode of \algo\ }\label{tab:pseudocode}
\end{algorithm*}

Let us now conclude this section by sketching the analysis of computational complexity of \algo.
Let $n$ be the input size (the length, in terms of words, of the ADR description). Let $m$ be the cardinality of the medical dictionary (i.e., the number of terms). Moreover, let $m'$ be the number of words occurring in the dictionary  and let $k$ be the length of the longest $t \in \llt$. For \med, we have around 70K terms and 20K words. Notice that $k$ is a very small constant. We assume that all update operations on auxiliary data structures require constant time.
Building  meta-dictionaries $\mathsf{DictByWord}$ and $\mathsf{DictByWordstems}$ requires $O(mk)$ time units. In fact, the simplest procedure to build the hash table is to scan the \llt\ dictionary and, for each term $t$, to verify for each word $w$ belonging to $t$ whether $w$ is a key in the hash table (this can be done in constant time). If $w$ is a key, then we have to update the values associated to $w$, i.e., we add $t$ to the set of terms containing $w$. Otherwise, we add the new  key $w$ and the associated term $t$ to the hash table.
Therefore, it can be easily verified that the voting procedure requires in the worst case $O(nm)$ steps, when a word belongs to all the \llt\ terms.
The computation of criteria-related weights requires $O(n)$ time units; the complexity of multi-value sorting can be approximated to  $O(n logn)$ time units (since the number of the criteria-related weights involved in the multi-sorting is fixed to be 5). Finally, deriving the best solutions actually requires $O(nl)$ steps.  

The computational complexity of \algo\ is likely to be lower than that of the tool proposed in~\cite{Uppsala}. Indeed, in \cite{Uppsala} the author describes a sophisticated procedure which considers also approximate string matching. This feature does not allow constant time search for text-dictionary  matches (i.e., it is not always possible to exploit direct data access through optimal data structures, such as hash tables). Moreover, explicitly considering word permutation and combination is a computationally  expensive task. We claim that the efficiency of \algo\ can be preserved also extending it with more advanced features, such as recognition of words in presence of orthographical errors. 
As a future work, we plan to provide formal and experimental comparisons of performances of \algo\ with respect to the software proposed in~\cite{Uppsala}.

\section{Software Implementation and Testing}\label{sec:uiandbenchmark}

\subsection{The User Interface}\label{sec:ui}

\begin{figure*} 
\begin{center}
\includegraphics[scale=.3]{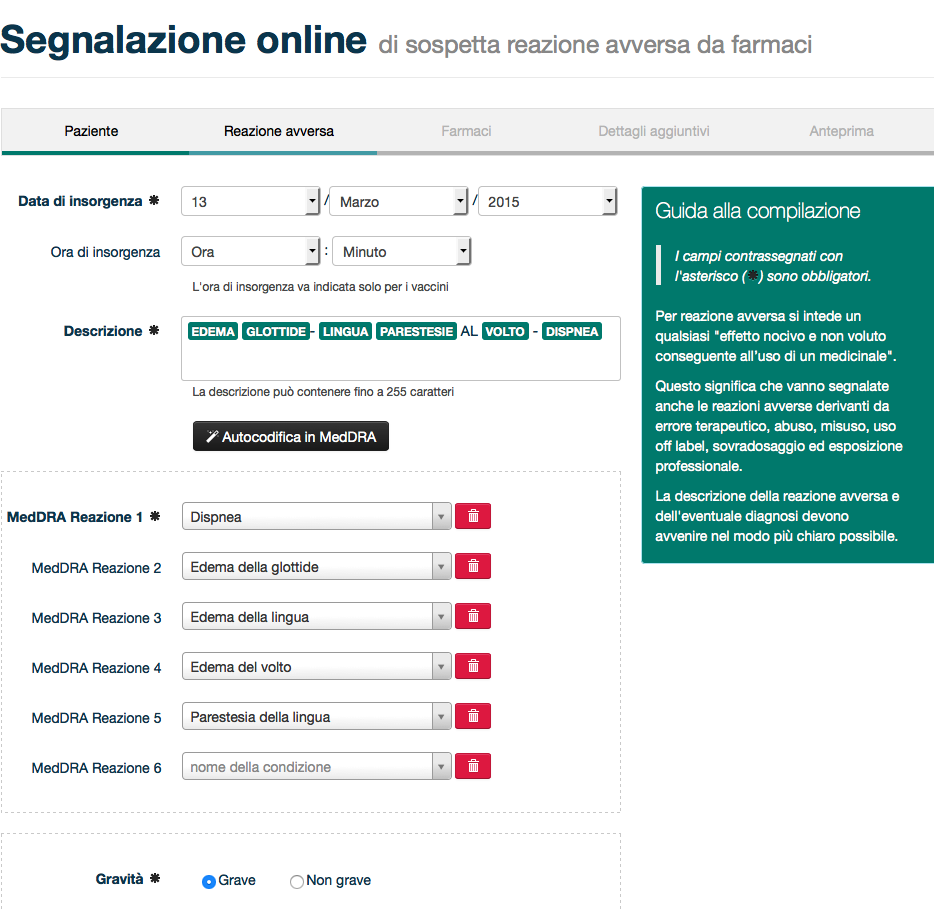}
\caption{A partial screenshot of \vigi\ User Interface}
\label{fig:vigiWScreen}
\end{center}
\end{figure*}

\algo\ has been implemented as  a \vigi\ plug-in: people responsible for pharmacovigilance  can  extract the auto-encoding of the narrative description and then revise and validate it.  
Figure~\ref{fig:vigiWScreen} shows a screenshot of \vigi\, for the part supporting back-end tasks (done by responsibles for pharmacovigilance revision activities). In the high part of the screen it is possible to observe the five sections composing a report. The screenshot actually shows the result of a human-\algo\ interaction: by pressing the button ``Autocodifica in \med'' (in English, ``\med\ autoencoding''), the responsible for pharmacovigilance obtains a \med\ encoding for the natural language ADR in the field ``Descrizione" (in English, ``Description''). 
Six solutions are proposed as the best \med\ term candidates: the responsible can refuse a term (through the trash icon), change one or more terms (by an option menu), or simply validate the automatic encoding and switch to the next section ``Farmaci'' (in English,  ``Drugs''). 
We are testing \algo\ performance in the daily pharmacovigilance activities. Preliminary qualitative results show that \algo\ drastically reduces the amount of work required for the revision of a report, allowing the pharmacovigilance stakeholders to provide high quality data about suspected ADRs.

\subsection{Testing}\label{sec:testing}


As a preliminary step in evaluating \algo\ performances, we developed a benchmark, which automatically compares \algo\ behavior with human encoding on already manually revised  and validated \adr\ reports.

To this end, we exploited  \vigis, a data warehouse and OLAP system for the italian pharmacovigilance activities developed for the Italian Pharmacovigilance National Center \cite{Vigisegn}. This system is based on the open source business intelligence suite Pentaho.  \vigis\ offers a large number of \emph{encoded}  \adr s. The encoding has been manually performed and validated by experts working at pharmacovigilance centres. Encoding results have then been sent to the national regulatory authority AIFA.

We performed a test, composed by the following steps. 

\begin{enumerate}
\item We launch an ETL procedure through Penthao Data Integrator. The procedure transfers reports from \vigis\  to an ad-hoc database \textsf{TestDB}. The dataset covers all the 6780 reports  received, revised, and validated during the year 2014 for the Italian region Veneto.
\item We launch an ETL procedure which extracts from reports stored in \textsf{TestDB} the narrative descriptions. For each description, the procedure calls \algo\ from \vigi; the output, i.e., a list of \med\ terms, is stored in a table of \textsf{TestDB}. 
\item Manual and automatic solutions, i.e., \llt\ term sets, are finally compared through an SQL query. We compute how much manual solutions are ``covered'' by \algo. In other words, we perform a similarity test between the two output sets. In order to have two uniform data sets, we map each \llt\ term, both from the manual and the automatic solutions, to its corresponding preferred term.
\end{enumerate}

Table~\ref{tab:percentage} shows the results of this first performance test.

\begin{table*}
  \begin{center}
    \begin{tabular}{| c| c |}
      \hline
      \textbf{Length of the Description (\# chars)} & \textbf{Percentage of global  identical solutions at the PT level}   \\
      \hline
         &\vspace{-.2cm}\\
      Short descriptions (up to 20 chars)& 81\% \\
          \hline
      &\vspace{-.2cm}\\
      Short/medium descriptions (from 20 up to 40 chars)& 62\%  \\
           \hline
      &\vspace{-.2cm}\\
      Medium descriptions (from 40 up to 100 chars) & 62\%\\
      \hline
        &\vspace{-.2cm}\\
      Long descriptions (from 100 up to 250 chars) & 61\%\\
           \hline
    \end{tabular}
  \end{center}
\caption{First results of \algo\ performances}\label{tab:percentage}
\end{table*}

It is worth noting that this test simply estimates how much \algo\ behavior is similar to the manual work on the \emph{whole set of solutions}, without considering the quality of  the manual encoding.  
We may observe that for short descriptions \algo\ results are very close to those from manual encoding. The percentage of similarity decreases with the growing of the number of characters, but it is stable beyond a certain threshold.  It could suggest that \algo\ will behave very well on very long (intractable) descriptions: as a human reviewer, the procedure does not encode redundant text.  
Since we did not evaluate the quality of the human solutions we take into account, we are working on a further quantitative analysis of \algo\ performances. We are developing an experimental test, involving three experts in report revision.  Two experts (and a third one, in case a reconciliation of diverging encoding is needed) are manually encoding a representative sample of ADR descriptions (about 200), in order to build a ground truth data set. These  ``certified'' manual solutions will be compared, report by report, with \algo's outputs. The test has been designed to effectively measure  soundness and  completeness of \algo. Informally, soundness can be estimated with respect to false positive terms provided by \algo; completeness can be estimated  according to \llt\ terms omitted by \algo. We will precisely quantify the difference between the human and the automatic encoding (taking into account also syntactically different but semantically equivalent solutions) and, thus, we will be able to compute the standard deviation of the behavior of the procedure w.r.t. the expected performance. 

\subsection{Examples}\label{sec:examples} 


Table~\ref{table:example} provides some examples of the behavior of \algo. We propose some free-text ADR descriptions from  \textsf{TestDB} and we provide both the manual and the automatic encodings into  \llt\ terms. We also provide the English translation of the natural language text (we actually provide a quite straightforward \emph{literal} translation).

\noindent 
{\scriptsize
D1--\texttt{~anaphylactic shock (hypotension + cutaneous rash) 1 hour after taking the drug.}\\
\noindent
D2--\texttt{~swelling in vaccination location left from 11/5; temperature less than 39,5 from 11/21; vesicles, blisters around the cheek from 11/10.}\\
\noindent
D3--\texttt{~extended local reaction, local pain, headache, fever for two days.}
}\\

\begin{table*}
    \begin{tabular}{|c|c|c|c|}
      \hline
 \# & \textbf{Narrative Description} & \textbf{LLT Human Encoding}  & \textbf{LLT \algo\ Encoding}   \\
      \hline
    D1 &   Shock anafilattico (\textit{ipotensione} + rash cutaneo)& $\underline{\mbox{Shock anafilattico}}^1$ & \textbf{Ipotensione}, $\underline{\mbox{Shock anafilattico}}^1$\\
     &  1 h dopo assunzione x os del farmaco &  &\\
          \hline
   D2 &  gonfiore in sede di vaccinazione sx dal 5/11,& $\underline{\mbox{Gonfiore in sede di vaccinazione}}^1$,  & \textbf{Bolle}, $\overline{\mbox{Febbre}}^2$, $\underline{\mbox{Gonfiore in sede di vaccinazione}}^1$, \\
   & febbre meno di 39,5 dal 21/11,     & $\overline{\mbox{Piressia}}^2$, $\overline{\mbox{Vescicole}}^3$ & $\overline{\mbox{Vescicole in sede di vaccinazione}}^3$\\
   &vescicole, \textit{bolle} presso la guancia dal 10/11&&\\
      \hline
    D3 & Reazione locale estesa, \textit{dolore} locale; & $\underline{\mbox{Cefalea}}^1$, $\underline{\mbox{Febbre}}^2$, & $\underline{\mbox{Cefalea}}^1$, \textbf{Dolore}, $\underline{\mbox{Febbre}}^2$, \\
  & cefalea e febbre per due giorni   &$\overline{\mbox{Reazione in sede di vaccinazione}}^3$ & $\overline{\mbox{Reazione locale}}^3$ \\
           \hline
    \end{tabular}
    \vspace{1ex}
\caption{Examples of \algo\ behavior}\label{table:example}
\end{table*}

In Table~\ref{table:example} we use the following notations: $\underline{t_1}^{n}$ and $\underline{t_2}^{n}$ are two \emph{identical} \llt\ terms retrieved both by  the human and the automatic encoding; $\overline{t_1}^{n}$ and $\overline{t_2}^{n}$ are two \emph{semantically equivalent} or \emph{similar} \llt\ terms retrieved  by  the human and the automatic encoding, respectively; we use bold type to denote terms  that are recognized by \algo\ and that have not been encoded by the reviewer; we use italic type in D1, D2, D3, to denote text recognized only by \algo. For example, in description D3, ``cefalea'' (in English, ``headache'')  is retrieved and codified both by the human reviewer and  \algo; in description D2,  ADR ``febbre'' (in English, ``fever') has been codified with the term itself by the algorithm, whereas the reviewer  codified it with its synonym ``piressia''; in D1,  ADR ``ipotensione'' (in English, ``hypotension''), has been retrieved only by \algo.

\section{Conclusions and Future Work}\label{sec:future}

In this paper we propose \algo, a simple and efficient NLP software, able to provide a concrete support to pharmacovigilance task, in the revision of ADR spontaneous reports. \algo\ takes in input a narrative description of a suspected ADR and produces as outcome a list of \med\ terms that ``cover'' the medical meaning of the free-text description.
We presented and implemented the first version of the algorithm, and preliminary results about its performances are encouraging. 

Finally, let us sketch here some ongoing and future work. First, we aim to prove that \algo\ is robust with respect to language (and dictionary) changes.  The way the algorithm has been developed suggests that \algo\ can be a valid tool also for narrative descriptions written in English. Indeed,  the algorithm retrieves a set of words, which covers an \llt\ term $t$, from a free-text description, without considering the order between words or the structure of the sentence. This way, we avoid the problem of ``specializing'' \algo\ for any given language. 
Furthermore, \algo\ performances can be strengthened, still maintaining the simple ``skeleton'' we proposed, eventually embedding new features \emph{inspired} to advanced NLP techniques. 
Even though negative sentences seem to be uncommon in ADR descriptions (at least in the data set we analyzed), the detection of negative forms is  a short-term issue we aim to address.
As a first step, we plan to recognize words that may represent negations and to signal them to the reviewer through the graphical UI. In this way, the software sends to the report reviewer an alert about the (possible) failure of the syntactical word-by-word recognition.
Moreover, we plan to address the  management of orthographical errors possibly contained in narrative ADR descriptions. We did not take into account this issue in the current version of \algo. A solution could be including an ad-hoc (medical term-oriented) spell checker in \vigi, to point out to the user that she/he is doing some error in writing the current word in the free description field. This should drastically reduce users' orthographical errors without  
{heavy side effects} in \algo\ development and performances.
As a further  extension of \algo, we will enrich the algorithm with heuristics and synonyms dictionaries. Moving towards the use of ad-hoc thesaurus dictionaries, our idea is to progressively (through everyday learning and feedback coming from experience) \emph{extend} \med\ with synonyms of \llt\ terms.
Finally, we aim to apply  \algo\ (and its refinements) to several different sources for ADR detection, such as, for example, drug information leaflets. 


\begin{thebibliography}{10}

\bibitem{SafetyDrug}
N.~Arthur, A.~{Bentsi-Enchill}, and {R. Couper et al.}, \emph{The Importance of
  Pharmacovigilance - Safety Monitoring of Medicinal Products.}\hskip 1em plus
  0.5em minus 0.4em\relax World Health Organization, 2002.

\bibitem{Borg11}
J.~Borg, G.~Aislaitner, M.~Pirozynski, and S.~Mifsud, ``Strengthening and
  rationalizing pharmacovigilance in the {EU}: where is europe heading to? a
  review of the new {EU} legislation on pharmacovigilance,'' \emph{Data
  Safety}, vol.~34, no.~3, pp. 187--197, 2011.

\bibitem{Aagard12}
L.~Aagaard, J.~Strandell, L.~Melskens, P.~Petersen, and E.~{Holme Hansen},
  ``Global patterns of adverse drug reactions over a decade: analyses of
  spontaneous reports to vigibase,'' \emph{Drug Safety}, vol.~35, pp.
  1171--1182, 2012.

\bibitem{Jurafsky2000}
D.~Jurafsky and J.~H. Martin, \emph{Speech and Language Processing: An
  Introduction to Natural Language Processing, Computational Linguistics, and
  Speech Recognition}, 1st~ed.\hskip 1em plus 0.5em minus 0.4em\relax Upper
  Saddle River, NJ, USA: Prentice Hall PTR, 2000.

\bibitem{Bate}
A.~Bate and S.~Evans, ``{Quantitative signal detection using spontaneous ADR
  reporting},'' \emph{Pharmacoepidemiology and Drug Safety}, vol.~18, no.~6,
  pp. 427--436, 2009.

\bibitem{Wang09}
X.~Wang, G.~Hripcsak, M.~Markatou, and C.Friedman, ``Active computerized
  pharmacovigilance using natural language processing, statistics, and
  electronic health records: {A} feasibility study,'' \emph{{JAMIA}}, vol.~16,
  no.~3, pp. 328--337, 2009.

\bibitem{Friedman}
C.~Friedman, ``Discovering novel adverse drug
  events using natural language processing and mining of the electronic health
  record,'' in \emph{Artificial Intelligence in
  Medicine}, ser. Lecture Notes in Computer Science.\hskip 1em plus 0.5em
  minus 0.4em\relax Springer Berlin Heidelberg, 2009, vol. 5651, pp. 1--5.

\bibitem{Aramaki10}
E.~Aramaki, Y.~Miura, M.~Tonoike, T.~Ohkuma, H.~Masuichi, and K.~Waki,
  ``Extraction of adverse drug effects from clinical records,'' \emph{Stud
  Health Technol Inform}, vol. 160, no. Pt1, pp. 739--43, 2012.

\bibitem{Dunagan08}
M.~G.~R. Reichley, P.~Kilbridge, L.~Noirot, R.~N. R, W.~Dunagan, and T.~Bailey,
  ``Natural language processing to identify adverse drug events,'' in
  \emph{{AMIA} Annu Symp Proc.}, 2008.

\bibitem{Bailey09}
P.~M. Kilbridge, L.~A. Noirot, R.~M. Reichley, and T.~C. Bailey, ``Computerized
  surveillance for adverse drug events in a pediatric hospital,'' \emph{J Am
  Med Inform Assoc.}, vol.~16, no.~5, pp. 607--612, 09.

\bibitem{Toldo12}
H.~Gurulingappa, A.~Mateen-Rajput, and L.~Toldo, ``Extraction of potential
  adverse drug events from medical case reports,'' \emph{Journal of Biomedical
  Semantics}, vol.~3, no.~15, pp. 1--10, 2012.

\bibitem{Gonzalez14}
A.~Sarker and G.~Gonzalez, ``Portable automatic text classification for adverse
  drug reaction detection via multi-corpus training,'' \emph{Journal of
  Biomedical Informatics}, vol.~53, pp. 196--207, 2015.

\bibitem{Yang12}
C.~C. Yang, H.~Yang, L.~Jiang, and M.~Zhang, ``Social media mining for drug
  safety signal detection,'' in \emph{Proc. of the 2012 Int. Workshop on Smart
  Health and Wellbeing, {SHB} 2012}, 2012, pp. 33--40.

\bibitem{Harpaz10}
R.~Harpaz, H.~S. Chase, and C.~Friedman, ``Mining multi-item drug adverse
  effect associations in spontaneous reporting systems,'' \emph{{BMC}
  Bioinformatics}, vol.~11, no. {S-9}, p.~S7, 2010.

\bibitem{Bellazzi15}
N.~Nissim, M.~Boland, R.~Moskovitch, N.~Tatonetti, Y.~Elovici, Y.~Shahar, and
  G.~Hripcsak, ``An active learning framework for efficient condition severity
  classification,'' in \emph{Artificial Intelligence in Medicine (AIME'15)},
  ser. Lecture Notes in Computer Science.\hskip 1em plus 0.5em minus
  0.4em\relax Springer, 2015, vol. 9105, pp. 13--24.

\bibitem{Uppsala}
G.~Dalhberg, ``{Implementation and evaluation of a text extraction tool for
  adverse drug reaction information},'' 2010, master Thesis, Uppsala University
  School of Engineering.

\bibitem{Collins03}
M.~Collins, ``Tutorial: Machine learning methods in natural language
  processing,'' in \emph{Computational Learning Theory and Kernel Machines,
  16th Annual Conference on Computational Learning Theory}, 2003, p. 655.

\bibitem{Radhakrishna14}
P.~Radhakrishna, ``Upversioning {MedDRA} dictionary - insights from a seasoned
  coder,'' \emph{Data Basics}, vol.~20, no.~3, pp. 1171--1182, 2014.

\bibitem{BauerLaurie2003}
L.~Bauer, ``Introducing linguistic morphology,'' 2003.

\bibitem{Kishida2005}
K.~Kishida, ``Technical issues of cross-language information retrieval: A
  review,'' \emph{Inf. Process. Manage.}, vol.~41, no.~3, pp. 433--455, May
  2005.

\bibitem{Clark2010}
A.~Clark, C.~Fox, and S.~Lappin, Eds., \emph{{The Handbook of Computational
  Linguistics and Natural Language Processing}}, ser. Blackwell Handbooks in
  Linguistics.\hskip 1em plus 0.5em minus 0.4em\relax John Wiley \& Sons, 2010.

\bibitem{string}
J.~Piskorski and M.~M.~Sydow, ``String distance metrics for reference matching
  and search query correction,'' in \emph{Business Information Systems}, ser.
  Lecture Notes in Computer Science, W.~Abramowicz, Ed.\hskip 1em plus 0.5em
  minus 0.4em\relax Springer Berlin Heidelberg, 2007, vol. 4439, pp. 353--365.

\bibitem{Vigisegn}
A.~Sabaini, ``Temporal data analysis and mining: A multidimensional approach
  and its application in a medical domain,'' Ph.D. dissertation, Department of
  Computer Science, University of Verona - Italy, 2015.

\end{thebibliography}

\end{document}